\documentclass[10pt,twocolumn,letterpaper]{article}

\usepackage{cvpr}
\usepackage{times}
\usepackage{epsfig}
\usepackage{graphicx}
\usepackage{amsmath}
\usepackage{amssymb}
\usepackage{amsfonts}
\usepackage{subfig}
\usepackage{authblk}

\usepackage[pagebackref=true,breaklinks=true,letterpaper=true,colorlinks,bookmarks=false]{hyperref}

\cvprfinalcopy 

\def\cvprPaperID{1337} 

\newcommand{\tb}[1]{{\textbf{#1}}}
\newcommand{\ti}[1]{{\textcolor{blue}{\textit{#1}}}}

\def\etal{et al\onedot}
\def\ie{\emph{i.e.}\xspace}
\def\eg{\emph{e.g.}\xspace}
\DeclareMathOperator*{\argmax}{arg\; max}

\ifcvprfinal\fi
\begin{document}



\title{Self-Supervised Video Representation Learning With Odd-One-Out Networks}


\author[1]{Basura~Fernando}
\author[2]{Hakan~Bilen}
\author[3]{Efstratios~Gavves}
\author[1]{Stephen~Gould}
\affil[1]{ACRV, The Australian National University, ACT 2601, Australia.}
\affil[2]{Visual Geometry Group, University of Oxford.}
\affil[3]{QUVA Lab, University of Amsterdam.}


\maketitle

\begin{abstract}
We propose a new self-supervised CNN pre-training technique based on a novel auxiliary task called \emph{odd-one-out learning}.
In this task, the machine is asked to identify the unrelated or \emph{odd} element from a set of otherwise related elements.
We apply this technique to self-supervised video representation learning where we sample subsequences from videos and ask the network to learn to predict the odd video subsequence. 
The odd video subsequence is sampled such that it has wrong temporal order of frames while the even ones have the correct temporal order.
Therefore, to generate a odd-one-out question no manual annotation is required.
Our learning machine is implemented as multi-stream convolutional neural network, which is learned end-to-end.
Using odd-one-out networks, we learn temporal representations for videos that generalizes to other related tasks such as action recognition.

On action classification, our method obtains 60.3\% on the UCF101 dataset using only UCF101 data for training which is approximately 10\% better than current state-of-the-art self-supervised learning methods.
Similarly, on HMDB51 dataset we outperform self-supervised state-of-the art methods by 12.7\% on action classification task. 
\end{abstract}

\section{Introduction}
\begin{figure}[t]
  \centering
  \includegraphics[width=0.99\linewidth]{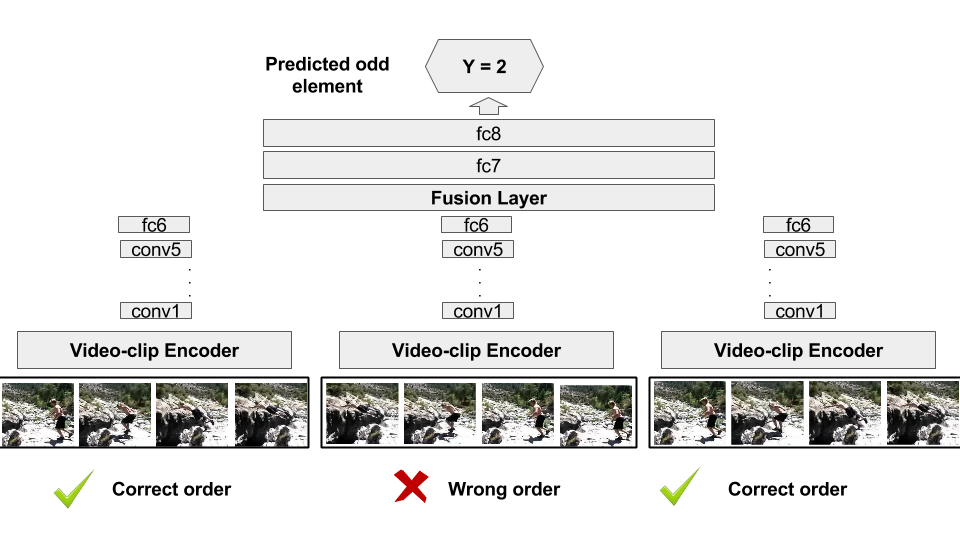}
  \caption{The proposed odd-one-out network, where it takes several video sequences as input to the multi branched network that share weights. 
  Objective is to identify the odd video sequence in this case it is the second video. To find the odd video-clip, learning machine has to compare all video clips, identify the regularities among them, and pick the one with irregularities. This type of tasks are know as \emph{analogical reasoning} tasks.} 
  \label{fig.iqq}  
\end{figure}
Convolutional Neural Networks (CNNs)~\cite{Lecun98} have emerged as the new state-of-the-art learning framework for many machine learning problems.
The success of CNNs has been largely sustained by the manual annotation of big datasets such as ImageNet~\cite{Russakovsky15} and Sports-1M~\cite{karpathy2014large}.
As manual annotations are costly and time consuming supervised learning becomes less appealing, especially when considering tasks involving more complex data (\eg, videos) and concepts (\eg, for human behavior analysis).
In this work we focus on learning video representations from unlabeled data.

Good video feature learning without using action category labels from videos is crucial for action recognition for two reasons.
First, unlike static images, videos are generally open-ended media and one cannot \emph{a priori} contain a particular action within a particular frame range.
Thus, for supervised learning one would need to manually annotate videos frame-by-frame or crop them to a range of frames to ensure consistency, obviously an unrealistic expectation for both.
Second, existing large video datasets, \eg, Sports-1M~\cite{karpathy2014large} and the recent YouTube-8M~\cite{AbuElHaija2016}, rely on noisy, unreliable YouTube tags.
As a result, one cannot truly identify whether it is the architecture or the noisy labels that contribute to the observed network behavior.
Besides, unlabeled videos are abundant and information rich regarding spatio-temporal structures~\cite{Bilen2016, Pickup14, Bengio09, Mobahi2009, Wiskott02}.

Although traditionally unsupervised feature learning (\eg \cite{Bourlard1988,Hinton1994}) implies no supervisory signals, recently researchers introduced the \emph{self-supervised} learning paradigm~\cite{Agrawal2015, Cruz2017, Doersch2015, Dosovitskiy2014, Wang2015}. Here the structure of the data is used as a supervisory signal so the method is unsupervised in the sense that it does not require human annotation but supervised machine learning techniques can still be used.
For instance, one can use relative location of the patches in images~\cite{Doersch2015} or the order of video frames~\cite{Wang2015} as a supervisory signal. 
Different from the above works we express supervision in the context of the \emph{odd-one-out} problem. More specifically, each training example is comprised of a \emph{question} composed of $N+1$ elements (such as $N+1$ video clips or images). Out of these $N+1$ elements $N$ are similar or related (\eg, correctly ordered set of frames coming from a video) and one is different or odd (\eg, wrongly ordered set of frames from a video). 
Both the odd element and the $N$ coherent elements are presented to the learning machine. 
The learning machine is then trained to predict the odd element.
To avoid a trivial solution, in each of the odd-one-out questions, the odd element is presented to the learning machine at random.
In particular, we use a CNN as the learning machine, a multi-branch neural network as illustrated in Figure~\ref{fig.iqq}.
During training, our method learns features that solves the odd-one-out problem.
As the network performs a reasoning task about the validity of elements (e.g. video subsequences), the learned features are useful for many other related yet different tasks.
Specifically, in this paper we demonstrate the advantages of odd-one-out networks to learn features in a self-supervised manner for video data.

Exploiting the spatio-temporal coherence in videos has been investigated before for unsupervised learning~\cite{Bengio09, Mobahi2009, Wiskott02}. 
However, with the exception of few works~\cite{Cadieu08,Srivastava2015}, the focus has been on appearance representations, perceiving videos as a collection of frames. 
There has also been unsupervised temporal feature encoding methods to capture the structure of videos for action classification~\cite{Fernando2016b,Fernando2015,Fernando2016,Li2016,Srivastava2015}.
In contrast, we focus on learning the motion patterns within videos. 
Inspired by the recently proposed family of motion representations~\cite{Bilen2016, Sun2015, Wang2016} that compress arbitrary length video sequences into fixed-dimensional tensors while maintaining their spatio-temporal structure, we propose a new video-segment level representation learning strategy for videos in self-supervised manner.

In this work we present an alternative methodology for learning video segment representations in an self-supervised manner.
Our contributions are threefold: First, we propose a novel learning task, \emph{odd-one-out learning}, for optimizing model parameters without relying on any manually collected annotations. 
Second, we present a neural network architecture suitable for odd-one-out learning. Third, our experimental results indicate that the trained networks learn accurate representations, outperforming considerably other recently proposed self-supervised learning paradigms for video data.

\section{Related work}
Unsupervised feature learning is well studied in the literature. The most common techniques studied include auto-encoders~\cite{Bourlard1988,Hinton1994}, restricted Boltzmann machines~\cite{Hinton1986}, convolutional deep belief networks~\cite{Lee2009}, LSTMs and recurrent neural networks~\cite{Srivastava2015}.

A recent emerging line of research for learning representations without manual annotations is self-supervised learning~\cite{Cruz2017, Doersch2015, Kumar2015, Wang2015}.
Self-supervised methods do not require manual annotations, instead they exploit the structure of the data to infer supervisory signals, which can then be used with robust and trustworthy supervised-like learning strategies.
In Doersch \etal~\cite{Doersch2015} spatial consistency of images are exploited as a context prediction task to learn image representations.
Video data are also used to learn image representations. 
%
%
For example, Wang \etal~\cite{Wang2015} generate pairs-of-patches from videos using tracking and use a Siamese triplet network to learn image representations such that the similarity between two matching patches should be larger than similarity between two random patches. 
The matched patches will have intraclass variability due to changes in illumination, occlusion, viewpoint, pose, and clutter.
However, tracking is not always reliable. As shown by Kumar \etal~\cite{Kumar2015}, training of such triplet networks is not straightforward with the need to estimate stable gradients for triplet-based losses.  
Agrawal \etal~\cite{Agrawal2015} exploit egomotion as a labelling process to learn representations where they show that egomotion is a useful supervisory signal when learning features. 
Similar to Wang \etal~\cite{Wang2015}, they also train a Siamese network to estimate egomotion from two image frames and compare it to the egomotion measured with odometry sensors. The resulting learned features are somewhat similar.

Another variant of unsupervised feature learning relies on exemplar CNNs~\cite{Dosovitskiy2014}.
Here each image is transformed using a large number of transformations and a CNN is trained to recognize instances of transformed images. 
A disadvantage of such an approach is that each image becomes a class, hence for a million images one trains a one-million class CNN. 
Moreover, the learned invariances depend on the types of transformations. 
However, the approach generate consistent labels, which is important for self-supervised learning based on CNNs.

The direction of time-flow (forward or backward) in videos was studied in an inspiring work by Pickup \etal~\cite{Pickup14}. The authors investigate various motion representations to learn the arrow of time. Unsupervised learning of sequence encoding for video data was proposed by Srivastava \etal~\cite{Srivastava2015}, where an LSTM encoder was used to learn unsupervised video encodings.
The LSTM is trained such that the encoding of the forward video is similar to the LSTM encoding of the reverse video.
However, this method requires a pre-trained network (with supervision) to extract frame level features and thus it is not a unsupervised feature learning method.

More recently, a CNN-based unsupervised representation learning method was presented in Misra \etal~\cite{Misra2016}. 
In that work, the learning task is to verify whether a sequence of frames from a video is presented in the correct order or not. This method has two shortcomings: i) the binary learning formulation results in a relatively easy learning problem, ii) despite having to determine the correct temporal ordering of frames, the method does not learn to encode temporal information but only spatial. In contrast, our method exploits the analogical reasoning over sequences and pose the feature learning problem as a $N+1$ way multi class classification problem which is much harder than the binary verification problem (see Fig.~\ref{fig.iqq}). Our method also able to learn temporal information taking the advantage of recent developments~\cite{Bilen2016, Sun2015, Wang2016} which leads to a superior performance for action recognition tasks.  

Most of the prior work in action recognition is dedicated to hand-crafted features~\cite{Herath2017} such as dense trajectory features~\cite{Fernando2016,JainCVPR2014,wang2013dense,wang2013action}.
Recently, supervised convolutional features obtain state-of-the-art performance either using very large video collections or using 3D convolutions~\cite{Tran2014} or by fine-tuning ImageNet pre-trained models~\cite{Bilen2016,Fernando2016a,LiArxiv16,Wang2016,Zha2015}. Our work differs from these in that we learn a video representation from self-supervision without using external information such as optical flow data or transfer filter weights from ImageNet pre-trained models or use effective cross modality pre-training as done in~\cite{Wang2016}.

\section{Odd-one-out learning}
\label{sec.iqnetworks}
 \begin{figure*}[t]
  \centering
  \includegraphics[width=0.49\linewidth]{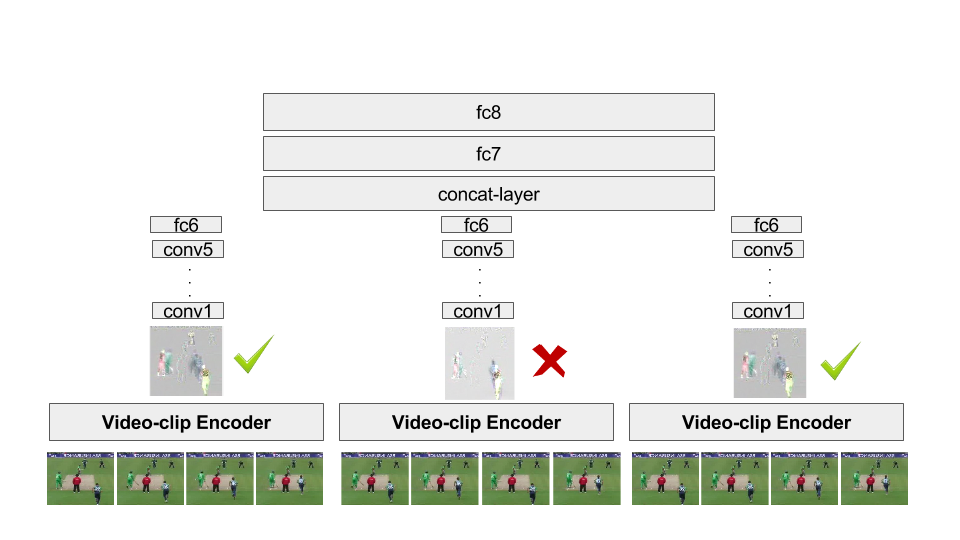}  
  \includegraphics[width=0.49\linewidth]{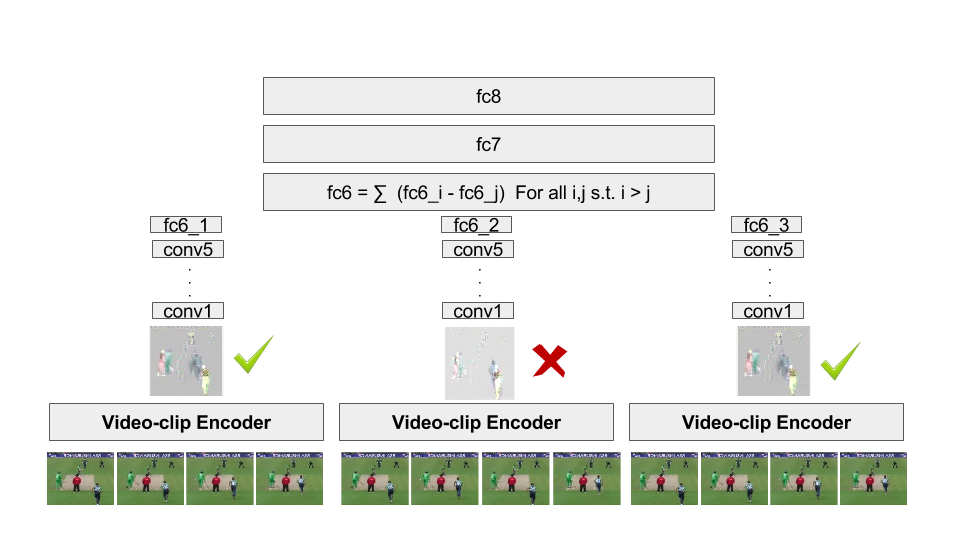} 
  \caption{The odd-one-out networks for video representation learning. Network is presented with one wrong sequence (out of order) and two correct sequences from the same video. Temporal encoder encodes the temporal structure of the subsequence. Odd-one-out network learn the feature to find out of order sequences. In the left figure, we see concatenation of FC6 activations and on the right we see the sum of difference network architecture. }
  \label{fig.video.odd.arh}
\end{figure*}

\noindent\textbf{Task}. The goal of odd-one-out learning is to predict the odd element from a set of otherwise related elements. 
There are different ways to generate such odd-one-out questions for both video or image data.
For example, in the case of video representation learning, the even objects could be correctly ordered video clips of a particular video, and the odd one could be a video clip obtained by wrongly permuting frames of the same video. 
This is just one example and our framework is quite general and can be applied to other data types, such as RGB image patches, video instances, or generic feature descriptors. 
The set of multiple related elements and the odd element comprise a question $q = \{I_1, \ldots, I_{N+1}\}$, where $I_i$ are the elements (in our case videos). 
We construct questions in an unsupervised manner. 
For example, in the context of feature learning for video classification, $I_1, \ldots, I_{N+1}$ are sets of sub-sequences sampled from a video. 
Out of these, $N$ number of sub-videos have the correct chronological order of frames which comprises the even set.
The odd video sub-sequence consist of frames sampled from an invalid order from the same video (see Figure~\ref{fig.iqq}). 
In both cases we know that one out of $(N+1)$ elements is an odd object.

In order to prevent a trivial solution,  we randomize the position of the odd element by a permutation $\sigma$ and obtain a question $q^\sigma$ with a respective answer $a^\sigma = \sigma(N + 1) \in \{1, 2, \ldots, N + 1\}$. 
The odd-one-out prediction task thus reduces to an $(N+1)$-way classification problem. 
Note, that given a set of unlabelled videos, we can automatically construct a self-supervised question-answer training set $\mathcal{D} = \{(q_j^{\sigma_j}, a_j^{\sigma_j})\}$, where the permutation $\sigma_j$ is chosen randomly for each question. 
Given this self-supervised dataset, the learning problem can be solved by standard maximum likelihood estimation, namely,
\begin{equation}
  \theta^\star = \argmax_{\theta} \mathcal{L}(f_\theta; \mathcal{D})
  \label{eq.subb.of.diff}
\end{equation}
where $\mathcal{L}$ is the likelihood function and $f_\theta$ is our parametrized model.

\noindent\textbf{Model}. We implement the prediction model $f_\theta$ as a multi-branch Convolutional neural network, which we call an \emph{odd-one-out network ($O3N$)}.
As illustrated in Figure~\ref{fig.iqq}, $O3N$ is composed of $N+1$ input branches, each contains five Convolutional layers and weights are shared across the input layers. 
Configuration of each input branch is identical to AlexNet architecture~\cite{Krizhevsky12} up to the first fully connected layer. As odd-one-out task requires a comparison among (N+1) elements of the given question and cannot be solved by only looking at individual elements, we introduce a fusion layer which merges the information from (N+1) branches after the first fully connected layer.
These fusion layers help the network to perform reasoning about elements in the question to find the odd one.
Ideally, the fusion layer should support the network to compare elements and find regularities and pick the element with irregularities.
We experiment with two fusion models, the \emph{Concatenation model} and \emph{sum of difference model} leading to two different network architectures as shown in Fig.~\ref{fig.video.odd.arh}.

\noindent\textbf{Concatenation model:} The first fully connected layers from each branch are concatenated to give a $(N+1) \times d$ dimensional vector, where $d$ is the dimensionality of the first fully connected layer.\\

\noindent\textbf{Sum of difference model:} The first fully connected layers from each branch are summed after taking the pair-wise activation difference leading to a $d$ dimensional vector, where $d$ is the dimensionality of the first fully connected layer. The advantage is that this strategy still encodes the structure of the odd-one-out feature activations yet can be represented with lower dimensional activation vector. Mathematically, let $\mathbf{v_i}$ be the activation vector of the $i$-th branch of the network. The output of the sum of difference layer is given by
\begin{equation}
 \mathbf{o} = \sum_{\forall j>i} \mathbf{v_j} - \mathbf{v_i}.
 \label{eq.sum.of.diff}
\end{equation}

We feed this fused activation vector through two fully connected layers followed by a softmax classifier with $N+1$ outputs.
Given a new training question $q^\sigma$, each input branch receives one of the $N+1$ elements and the network must learn to predict the location of the right answer, $a^\sigma$.
We illustrate in Figure~\ref{fig.iqq} our proposed $O3N$ together with an example question.

\section{Learning video representations with O3N}
\label{sec.video}
\begin{figure}[t]
  \centering
  \includegraphics[width=0.98\linewidth]{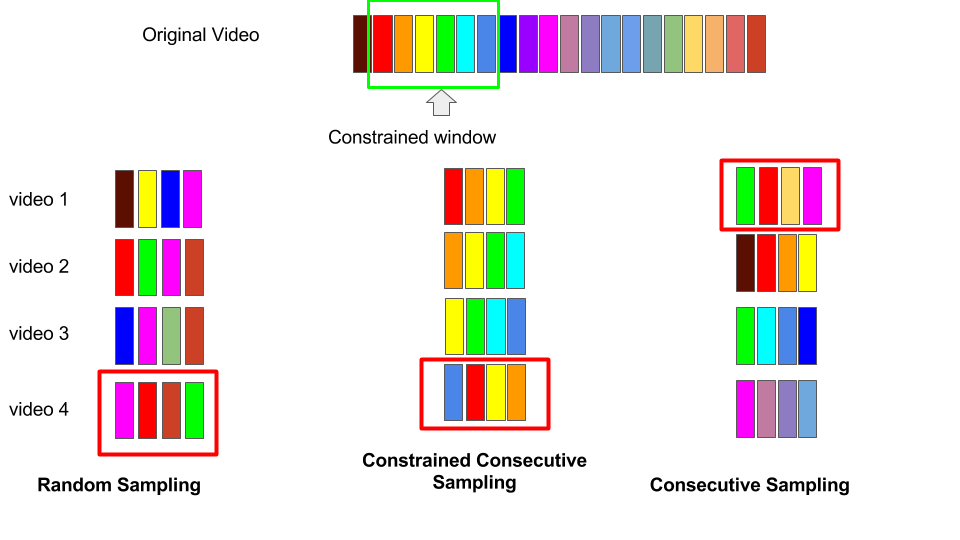}    
  \caption{Three different types of sampling strategies are investigated for odd-one-out learning. The red box shows the odd video sub-sequence from each of the sampling type. The constrained consecutive sampling sample from a constrained part of the original video which is denoted by the green box. The constrained sampling window is $1.5 \times W$ where $W$ is the length of the sampled sub-sequences (must be viewed in colour).}
  \label{fig.sampling}
\end{figure}

In this section we present a method to learn video representations in self-supervised manner using odd-one-out learning. 
In odd-one-out learning, we have to decide how to generate questions.
Mainly, our objective is not only to solve the odd-one-out problem, but also to learn good features.
If the odd-one-out task is generalizable and the generated questions are related to solving other related tasks, one can expect to obtain good representations for the input video data.

Our aim is to learn features that are useful for video classification.
Specifically, we are interested in action recognition from video data.
It is important to learn good temporal representations to solve the action recognition problem.
As videos are essentially composed of sequences of frames, by nature most of the videos have a strong temporal structure.
Hence, a good video representation should be able to capture this temporal structure within a sequence of frames.
We propose to employ the odd-one-out self-supervised learning  to exploit the structure within video sequences.
Therefore, we generate odd-one-out questions by exploiting the structure of the videos.

Specifically, let us assume we are given a video sequence $\mathcal{X} = \left< X_1, X_2, \cdots X_n \right>$ which consist of $n$ number of RGB frames.
The $t$-th RGB frame is denoted by $X_t$. 
Because videos are sequences, there are order constraints over frames such that  $X_1 \succ X_2 \succ \cdots \succ X_n$. 
The general idea for generating odd-one-out questions is to sub-sample $W$ frames from $\mathcal{X}$ where $W<n$. 
Then we generate elements of the odd-one-out questions by different sampling strategies. 
Each of these sampling strategy has implications on the learned features.
Next we discuss three sampling strategies, the consecutive sampling, random sampling, and constrained consecutive sampling. 

\noindent\textbf{Consecutive sampling:} 
We sample $W$ number of consecutive frames $N$ times from video $\mathcal{X}$ to generate $N$ number of even (related) elements. 
Each sampled even element of the odd-one-out question is a valid video sub-clip consisting of $W$ consecutive frames from the original video. 
However, the odd video sequence of length $W$ is constructed by random ordering of frames and therefore does not satisfy the order constraints. 
These random frames could come from any location of the original video (see figure~\ref{fig.sampling} right).
Then the objective of the odd-one-out video network is to learn to recognize the odd (wrong video sequence) out of other $N$ correct sequences.

\noindent\textbf{Random sampling:} 
We randomly sample $W$ frames $N$ times from the video $\mathcal{X}$ to generate $N$ number of even (related) elements. 
Each of these $N$ elements are sequences that has the correct temporal order and satisfy the original order constraints of $\mathcal{X}$. 
However, the frames are not consecutive as in the case of consecutive sampling.
The odd video sequence of length $W$ is also constructed by randomly sampling frames. 
An illustration is shown in Figure~\ref{fig.sampling} middle.
Similar to consecutive sampling strategy, the odd sequence does not satisfy the order constraints. 
Specifically, we randomly shuffled the frames of the odd element (sequence).

\noindent\textbf{Constrained consecutive sampling:} 
In the constrained consecutive sampling strategy, first we sub select a video clip of size $1.5 \times W$ from the original video which we denote by $\mathcal{\hat{X}}$. 
We randomly sample $W$ consecutive frames $N$ times from $\mathcal{\hat{X}}$ to generate $N$ number of even (related) elements. 
Each of these $N$ elements are subsequences that have the correct temporal order and satisfy the original order constraints of $\mathcal{X}$. 
At the same time each of the sampled even video clips of size $W$ overlaps more than 50\% with each other. 
The odd video sequence of length $W$ is also constructed by randomly sampling frames from $\mathcal{\hat{X}}$. 
Similar to other sampling strategies, the odd sequence does not satisfy the order constraints. 
Specifically, we randomly shuffled the frames of the odd element (sequence).

\section{Video frame encoding}
\label{sec.dynamic}
\begin{figure}[t]
\begin{center}
  \subfloat[][]{\includegraphics[width=0.20\linewidth]{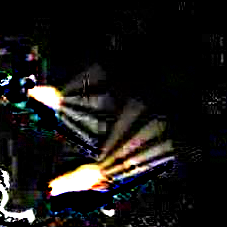}}   
  \subfloat[][]{\includegraphics[width=0.20\linewidth]{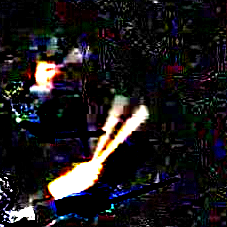}}\\
  
  \subfloat[][]{\includegraphics[width=0.19\linewidth]{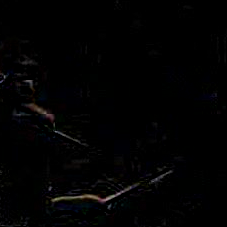}}
  \subfloat[][]{\includegraphics[width=0.19\linewidth]{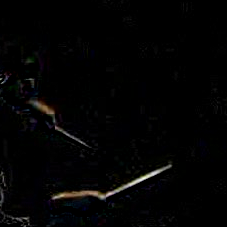}}  
  \subfloat[][]{\includegraphics[width=0.19\linewidth]{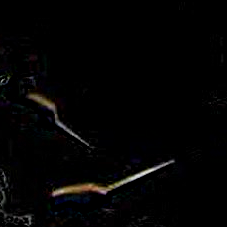}}
  \subfloat[][]{\includegraphics[width=0.19\linewidth]{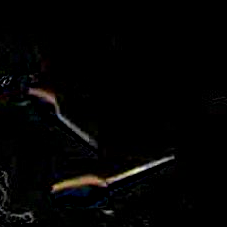}}
  \subfloat[][]{\includegraphics[width=0.19\linewidth]{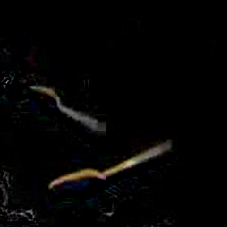}}  
\end{center}
\caption{Several video-clip encoder outputs for action drumming. (a) Dynamic image (b) sum of difference of frames (c-g) stack of difference of frames. All method use sequence size of 6 frames.}
\label{fig:dynamics}
\end{figure}

In this section we describe the video-clip encoding step of our odd-one-out network architecture. 
As shown in Fig.~\ref{fig.iqq}, each element (video-clip or subsequence) in an odd-one-out question is encoded to extract temporal information before presenting to the first convolutional filters of the network.
As mentioned in section~\ref{sec.video}, odd-one-out networks are presented with sub-sequences of videos. 
These sub-videos can be valid or invalid (wrong) video clips.
We want to use odd-one-out networks to learn video representations by exploiting the structure of the sequences.
There are several ways to capture the temporal structure of a video sequence.
For example, one can use 3D-convolutions~\cite{Ji2013}, recurrent encoders~\cite{Sutskever2014}, rank-pooling encoders~\cite{Fernando2016} or simply concatenate frames.
Odd-one-out networks can use any of the above methods~\cite{Ji2013,Sutskever2014,Fernando2016} to learn video representations in self-supervised manner using video data.

A single RGB image usually contains only static appearance at a specific time point and lacks the contextual information about previous and next frames. 
In contrast, the RGB difference between two consecutive frames describe the appearance change, which may correspond to the motion salient region. 
This information is also related to the velocity of the RGB data. 
Next, we discuss three technique that is used in our experiments to encode video-frame-clips using the differences of RGB frames into a single tensor $\mathcal{X}_d$.

\noindent
\textbf{Sum of differences of frames video-clip encoder:}
In this method we take the difference of frames and then sum the differences to obtain a single image $\mathcal{X}_d$.
This single image captures the structure of the sequence. 
Precisely, this is exactly same as the equation~\ref{eq.sum.of.diff} but now applied over frames instead of vectors.
It is interesting to note that this equation boils down to a weighted average of frames such that $\mathcal{X}_d = \sum w_t  X_t$ where the weight of frame at index $t$ is given by 
\begin{equation}
w_t = W + 1 - 2t. 
\end{equation}
If the input sequence has spatial resolution of $h \times w$ and temporal extent of $W$, then, the output image has the same spatial resolution but the temporal information is summarized into a single image of size $h \times w \times 3$ for R,G,B channels (see Fig.~\ref{fig:dynamics} (b)). 

\noindent
\textbf{Dynamic image~\cite{Bilen2016,Bilen2016a} encoder:} This method is similar to the \emph{sum of differences of frames} method, however the only difference is that now the input sequence is pre-processed to obtain a smoothed sequence $\mathcal{M}=\left< M_1, M_2, \cdots M_W \right>$. Smoothing is obtain using the mean at index $t$. The smoothed frame at index $t$ denoted by $M_t$ is given by 
\begin{equation}
M_t = \frac{1}{t}\sum_{j=1}^{t} X_j 
\end{equation} 
where $X_j$ is the frame at index $j$ of the sub-video. The dynamic image can be computed very efficiently. 
In-fact, dynamic image can be computer as a weighted linear combination of original frames where the weight at index $t$ is computed by $
w_t = 2(W - t +1) - (W+1) (H_T - H_{t-1})$.
Here $H_t = \sum_{i=1}^t \frac{1}{t}$ is the $t$-th Harmonic number and $H_0=0$. For complete derivation of Dynamic image we refer the reader to~\cite{Bilen2016,Bilen2016a}. 
An example of a Dynamic Image is shown in (see Fig.~\ref{fig:dynamics} (a)).

\noindent
\textbf{Stack of differences of frames video-clip encoder:} Inspired by~\cite{Sun2015,Wang2016}, we also stack the difference of frames instead of summing them. Once again the objective is to capture the motion and dynamics of short video clips. However, now the resulting image is not any more a standard RGB image with three channels. Instead, we obtain $ (N-1) \times 3$ channel image ((see the stack in Fig.~\ref{fig:dynamics} (c-g)). 

\section{Experiments}
In this section we explain the experimental set up and the experimental results which validate the effectiveness of our odd-one ($O3N$) learning. 
We evaluate the usefulness of our odd-one-out learned features on the action classification task. 
Specifically, we use UCF101 and HMDB51 datasets for self-supervised feature learning from video data and then use the features for action classification.

The \textit{UCF101 dataset~\cite{soomro2012ucf101}} is an action recognition dataset of realistic action videos, collected from YouTube, consists of 101 action categories. It has 13,320 videos from 101 diverse action categories. The videos of this dataset is challenging which contains large variations in camera motion, object appearance and pose, object scale, viewpoint, cluttered background and illumination conditions. It consist of three splits, in which we report the classification performance over all three splits as done in the literature.

The \textit{HMDB51 dataset~\cite{Kuehne2011}} is a generic action classification dataset consists of 6,766 video clips divided into 51 action classes. Videos and actions of this dataset are challenging due to various kinds of camera motions, viewpoints, video quality and occlusions. Following the literature, we use a one-vs-all multi-class classification strategy and report the mean classification accuracy over three standard splits provided by Kuehne~\etal~\cite{Kuehne2011}.

In rest of the sections, we perform several experiments to demonstrate different aspects of odd-one-out learning, network design choices, and the performance of different video-clip encoders when used with $O3N$ networks. 
\subsection{Default odd-one-out training for videos.}
In this section, we explain the default odd-one-out learning process.
By default, we use questions that consist of six video sequences, five sequences with frames in the correct order and a sequence with frames in a wrong order. 
Each sampled video subsequence consists of six frames sampled according the sampling process described in section~\ref{sec.video}. 
Unless otherwise specified we use the random sampling as the default sampling process. 
We rely on the AlexNet architecture, however, the number of activations in the first fully connected layer is reduced to 128 unless otherwise specified. 
The \emph{sum of difference model} architecture (see section~\ref{sec.iqnetworks}) is our default activation fusion method. 
By default, we use the Dynamic Image~\cite{Bilen2016} as the temporal video-clip encoder. 
Experiments are run for 200 epochs without batch normalization and with a learning rate varying from 0.01 to 0.0001 in logarithmic manner. The batches are composed of 64 questions. 
Each question consist of six sub-videos and each sub-video has six frames. 
The self-supervised network is trained with stochastic gradient descent using MatConvNet~\cite{Vedaldi2015}. We use the first split of UCF101 datasets for training of the odd-one-out networks and also for validation. 
Temporal jittering is used to avoid over-fitting.
\subsection{Fine tuning for action recognition.}
Once we train the odd-one-out network, with default setting, we use that to initialize the supervised training. 
We initialize the fine-tuning network (AlexNet architecture~\cite{Donahue13} with standard 4096 activation at fully connected layers) with the convolutional filter weights obtained from the odd-one-out network. 
The fully connected layers are fine-tuned with a learning rate 10 times larger than the ones used for convolutional layers ($10^{-2}$ to $10^{-4}$) and batches composed of 128 samples. 
Typically, the network takes sub-sequences of length six (six frames) as input (same size used in the the odd-one-out network).
We use temporal jittering and drop out rate of 0.8. 

During final inference, to compute the classification accuracy, we sample all non-overlapping sub-sequences (consists of six frames) and compute the maximum conditional probabilistic estimate per sequence. Mathematically, let us assume that given long video $\mathcal{X}$ we have sub-sample $m$ subsequences of size $W$, denoted by $\{  \hat{\mathcal{X}_i}  \}$ where $i=1 \cdots m$. 
Therefore the CNN returns the conditional probability of action category $y$ for subsequence $\hat{\mathcal{X}_i}$ which is denoted by $p(y|\hat{\mathcal{X}_i})$.  
During the final inference, using i.i.d. assumption, the conditional log probability of the class $y$ given video $\mathcal{X}$ is obtained by $\sum_{i=1}^{m} log(p(y|\hat{\mathcal{X}_i}))$. 
We use the category that returns the maximum log conditional probability as the predicted class for that video.
\subsection{Evaluating sampling types for $O3N$ learning.}
\begin{table}[t]
\small
\begin{center}
\begin{tabular}{|l|c|c|c|c|c|}
\hline
Method 				&  superv. acc.(\%) &  self.sup. acc.(\%)\\ \hline \hline
Random initialization 		&  47.0 		& n/a \\ \hline
$O3N$-consec. samp. 		&  50.6 		& 27.4\\ \hline
$O3N$-const. consec. samp. 	&  52.4 		& 29.0\\ \hline
$O3N$-random sampling 		&  \tb{53.2} 		& \tb{29.6}\\ \hline


\end{tabular}
\end{center}
\caption{Comparing several odd-one-out sampling strategies with the random initialization for video action classification on UCF101 dataset.}
\label{tbl.sampling.di}
\end{table}
The objective of the first experiment is to evaluate the impact of sampling types used for odd-one-out networks. 
In this experiment, we use the default setting for odd-one-out ($O3N$) training for videos and use the default fine-tuning process explained earlier. 
Odd-one-out training is performed only on the training set of the UCF101 datasets first split.
Learned features are used to fine-tune all three splits of the UCF101 separately to evaluate the action classification accuracy.
We compare the three sampling types explained in section~\ref{sec.video}, namely \emph{a)} consecutive sampling, \emph{b)} random sampling and \emph{c)} constrained consecutive sampling.
We also compare our $O3N$ initialization with the randomly initialized fine-tuning network results.
Results are reported in Table~\ref{tbl.sampling.di}.

As it can be seen from the results obtained in Table~\ref{tbl.sampling.di}, all three initialization methods that uses odd-one-out learning perform better than random initialization on the supervised action classification task. 
Random initialization obtains only 47.0\% over three splits where as $O3N$ consecutive sampling obtains 50.6\% which is 3.6\% better than random initialization. 
Interestingly, the \emph{constrained consecutive sampling} process obtains better results compared to \emph{consecutive sampling} (52.4\%). 
The \emph{random sapling} process obtains the best results for both supervised and self-supervused tasks. 

The consecutive sampling is more confusing task and therefore the most difficult for the network to solve.
Videos typically have slow motions and in that case it may be difficult to tell apart correct vs incorrect ordering using consecutive sampling.  
With such a confusing task, the network might learn little. Moreover, learning from small motions means focusing on small subtelties and this may not generalize well to large general motion understading. This might be the reason for the poor performance of constrained consecutive sampling compared to random sampling.
After analysing these results, we conclude that odd-one-out learning obtains better features that potentially generalizes for other tasks and applications such as video action classification. 
Secondly, more general $O3N$ tasks based on random question generations process such as \emph{random sampling} seems to generate more generalizable features.
Therefore, the odd-one-out learning does not need a carefully designed sampling process to learn good video features as apposed to methods such as~\cite{Misra2016}.
\subsection{Capacity of fully connected layers.}
We hypothesize that analogical reasoning tasks such as $O3N$ learning generates useful features. 
However, if one wants to capture such information in the convolutional filters, perhaps it is better to limit the capacity of the fully connected layers. 
To attain this objective, we have introduced two design choices. 
First, we have reduced the number of activations at the fully connected layers to have only 128 instead of 4096.
Secondly, we use sum of difference (SOD) as the fusion method instead of simply concatenating (CON) the activations in our multi-branch network architecture. 
In this experiment, we evaluate the impact of both these design choices. 
We use the default experiment protocol but now use $O3N$ with random sampling. 
First, we evaluate the impact of using 128 dimensional activations compared to 4096 using sum of difference model as the fusion method. Results are reported in Table~\ref{tbl.capacity.di}.
\begin{table}[t]
\scriptsize
\begin{center}
\begin{tabular}{|l|c|c|c|c|c|}
\hline
Method 		& self-sup. acc. & split- 1 & split 2 & split 3 & mean\\ \hline \hline
$O3N$-4096-SOD 	& 25.7	&51.7     & 51.4    & 50.9    & 51.3 \\ \hline
$O3N$-128-SOD	& 29.6	&\tb{54.1}     & \tb{51.9}    & \tb{53.6}    & \tb{53.2} \\ \hline 
$O3N$-128-CON 	& \tb{33.6}	&49.7     & 50.3    & 50.4    & 50.1 \\ \hline
\end{tabular}
\end{center}
\caption{Comparing the impact of capacity of $O3N$ networks fully connected layers on feature learning and action classification.}
\label{tbl.capacity.di}
\end{table}
Interestingly, a reduced capacity of 128 activations obtains better results than 4096 dimensional activations for both supervised learning and self-supervised learning. 
When the number of activations is reduced to 128, the self-supervised performance increase from 25.7\% to 29.6\% which is also reflected in supervised task where the supervised action classification performance improve from 51.3\% to 53.2\%.
It is also possible that this is partially, due to lack of over-fitting. 

Secondly, we compare the impact of multi branch fusion using feature concatenation (CON) with the sum of differences (SOD) fusion. 
Results are also reported in Table~\ref{tbl.capacity.di}. 
Now we are comparing $O3N$-128-SOD with $O3N$-128-CON.
Interestingly, the feature concatenation obtains the good results compared to sum of difference model for the self-supervised task.
However, the supervised action classification results for CON is not as good as the sum of difference (SOD) method. 
Even if sum of difference method (128-SOD) has relatively poor performance on the self-supervised task (29.6 compared to 33.6), intuitively it has the ability to push down the abstractions about analogical reasoning to the convolutional filters.
Therefore, the sum of difference model learns a better \emph{feature representations} in the expense of slight performance degradation on the task that it solves when used with odd-one-out learning.
\subsection{How big the $O3N$ questions should be?}
In this experiment we evaluate the impact of $O3N$ learning using different number of elements in each $O3N$ question. We use the default experimental protocol with random sampling and train network with 2, 4, 6, 8 and 10 elements (subsequences) in each of the question and report the supervised and unsupervised performance on the validation set. Note that the self-supervised task is only trained on the split 1 of UCF101 dataset. Self-supervised task evaluated on the validation set of UCF101 split 1. Results are reported on Table~\ref{tbl.num.questions}. Note that $O3N$ method with two elements reduces to what is similar to sequence verification method~\cite{Misra2016}.
\begin{table}[t]
\scriptsize
\begin{center}
\begin{tabular}{|l|c|c|c|c|c|}
\hline
Nq. 	& self.sup.acc. 	& split 1 & split 2 & split 3 & mean\\ \hline \hline
2	& \tb{73.0} 	& 49.3   	& 49.3    	& 49.4    & 49.3  	\\ \hline 
4	& 43.6 		& 52.1   	& 51.5    	& 51.3    & 51.6 	\\ \hline 
6	& 29.6 		& 54.1   	& 51.9    	& \tb{53.6}    & \tb{53.2} 	\\ \hline
8	& 21.3 		& \tb{54.5}   	& 52.5    	& 52.3    & 53.1	\\ \hline
10	& 16.6 		& 52.6   	& \tb{52.7}     & 53.2    & 52.8 	\\ \hline 
\end{tabular}
\end{center}
\caption{Impact of number of question (Nq.) on $O3N$ learning on UCF101 dataset.}
\label{tbl.num.questions}
\end{table}
As it can be seen from the results in Table~\ref{tbl.num.questions}, as we increase the number of elements in the $O3N$ question, the unsupervised task becomes harder. As a result, the unsupervised classification accuracy decreases. However, it is interesting to see that $O3N$ task with two elements obtains only 49.3\% accuracy on the supervised classification task. However, with the increment of elements in each question, it tends to obtain better results for supervised classification task. On average best supervised results are obtained for a $O3N$ question consist of six elements (\ie five related correct subsequences and one odd wrong subsequence). Results suggests as the task becomes very difficult (8 and 10 elements), the supervised results saturate and starts to decrease. 
This is because when tackling a very ambiguousand hard tasks the network may learn very little because it is not able to solve it which is also reflected in the poor performance.
When the task is too easy to solve, the network might also not learn much ( question size of 2). In Table~\ref{tbl.num.questions}, we see this effect.
\subsection{Video-clip encoding methods.}
In this section we compare the impact of $O3N$ learning using three video-clip encoding methods discussed in the section~\ref{sec.dynamic}. 
We evaluate the sum of difference of frames (Sum-of-diff.) video-clip encoding, with the Dynamic Image~\cite{Bilen2016} encoding, and  
the stacking of sum of difference of frames (Stck.-of-diff.) video-clip encoder.
We compare the results for action recognition using UCF101 and HMDB51 datasets on Table~\ref{tbl.dynamic}.

For UCF101 (Table~\ref{tbl.dynamic}) the random initialization using the (Sum-of-diff.) video-clip encoding method (in section~\ref{sec.dynamic}) obtains only 43.4\% while for the same video-clip encoder, the $O3N$ initialization obtains 54.3\% which is a significant improvement of 10.9\%.
Random initialization of dynamic images obtains better results than the sum of difference random initialization.
However, with $O3N$ learning, the obtained results for dynamic images is 1.1\% worse than the sum of difference method.
Most interestingly, the stacking of difference of frames obtains the best results for both random initialization and $O3N$ initialization.
Using $O3N$ learning we improve the random initialization results for all three video-clip encoder methods with improvements of 10.9\%, 6.2\% and 9.8\% indicating the advantage of $O3N$ learning for video representation learning. Similar trend can be seen for HMDB51 dataset as well (see Table~\ref{tbl.dynamic}). 
When the network is intialized with ImageNet pretrained models, we obtain 64.9 \%, 67.2 \%, 70.1 \% for sum-of-difference video-clip encoding, dynamic images and stack of difference methods respectively on UCF101.
\begin{table}[t]
\scriptsize
\begin{center}
\begin{tabular}{|l|c|c|}
\hline
Method 				& UCF101 & HMDB51 \\ \hline \hline
Rand Init. - Sum-of-diff.	& 43.4 & 21.8\\ \hline
$O3N$ - Sum-of-diff.		& \ti{54.3} & \ti{25.9}\\ \hline \hline
Rand Init. - Dynamic image.	& 47.0 & 22.3\\ \hline
$O3N$ - Dynamic image 		& \ti{53.2} & \ti{26.0}\\ \hline  \hline
Rand Init. - Stck.-of-diff. 	& 50.2 & 28.5 \\ \hline 
$O3N$ - Stck.-of-diff. 		&  \tb{60.0} & \tb{32.4}\\ \hline
\end{tabular}
\end{center}
\caption{Impact of several video-clip encoder methods for odd-one-out learning using UCF101 dataset and HMDB51 over three splits.}
\label{tbl.dynamic}
\end{table}
\subsection{Visualizing the learned network filters}
\begin{figure*}[t]
  \centering
  \includegraphics[width=0.30\linewidth]{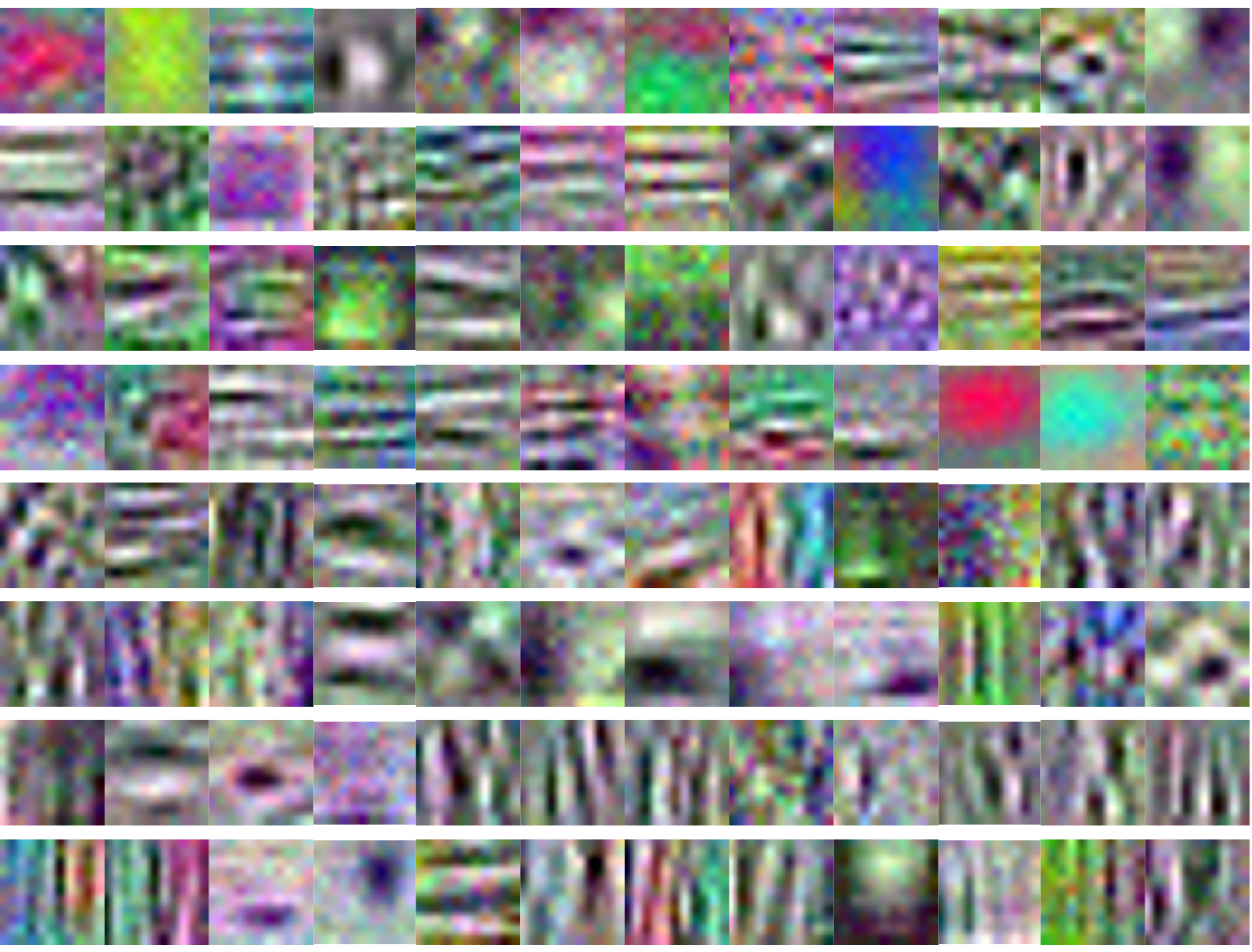}  
  \includegraphics[width=0.30\linewidth]{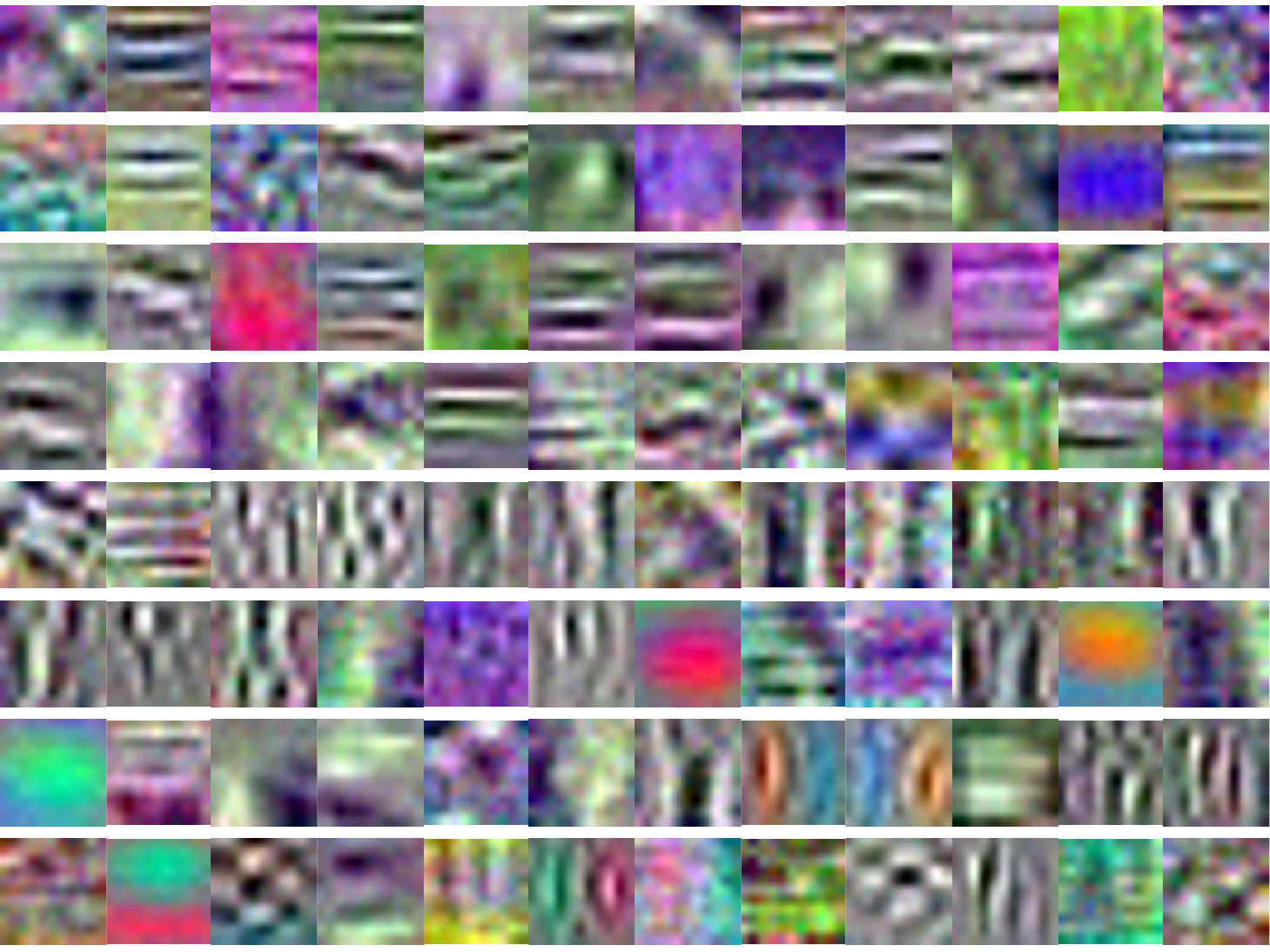} 
  \includegraphics[width=0.30\linewidth]{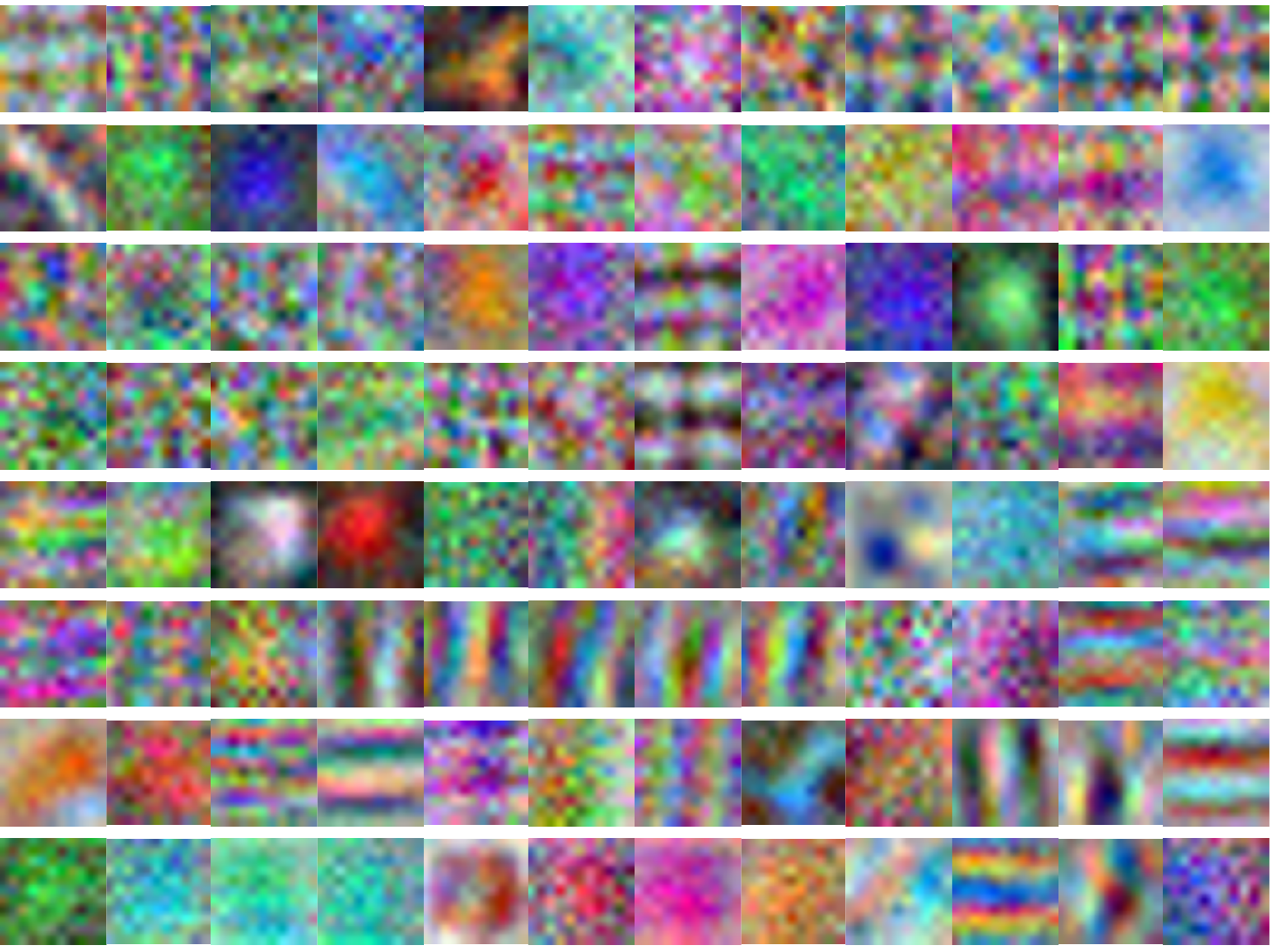}
  \caption{Visualizing the learned first convolutional filter weights for (a) sum of difference video-clip encoder (b) Dynamic Image video-clip encoder (c) Stack of difference video-clip encoder.}
  \label{fig:filters}
\end{figure*}

In this section we visualize some of the network filters learned from sum of difference video-clip encoder (Fig.~\ref{fig:filters}(a)), Dynamic Image video-clip encoder (Fig.~\ref{fig:filters}(b)) and stack of difference of frames video-clip encoder (Fig.~\ref{fig:filters}(c)). It is not surprising that the learned filters for sum of difference video-clip encoder (Fig.~\ref{fig:filters}(a)) and Dynamic Image video-clip encoder (Fig.~\ref{fig:filters}(b)) is some what similar. Interestingly, the stack of difference of frames video-clip encoder (Fig.~\ref{fig:filters}(c)) has totally different set of filters (note we are visualizing only the mean of the first 5 depths of the conv. filters as the Red channel, the next 5 depths of the filters as Green channel and the last 5 as the Blue channel). All filters, are obtained using six frame clips and using six elements per question which is the default 03N setting.
\subsection{Comparing with state-of-the-art.}
\label{sec.soa}
In this section, we compare our $O3N$-based self supervised results with the other state-of-the-art self-supervised methods. Specifically, we compare with DrLim~\cite{Hadsell2006}, TempCoh~\cite{Mobahi2009}, Obj. Patch~\cite{Wang2015} and Seq.Ver~\cite{Misra2016}. 
Results are reported in Table~\ref{tbl.soa}. Note that we use only the split 1 of UCF101 and HMDB51 so that we can compare with other published results~\cite{Misra2016}.
As it can be seen from the results, our $O3N$ learning-based features obtains score almost 10\% higher in UCF101 than the second best method reported in the literature~\cite{Misra2016} that relies on sequential verification. Similarly, we obtain massive improvement of 12.7\% for HMDB51 dataset over~\cite{Misra2016}.

It should be noted that when relying on deep architectures pretrained on supervised datasets, like Imagenet~\cite{Deng2009}, the state-of-the art reaches about 94.2\% (\cite{Wang2016}) using optical flow, improved trajectory features and RGB data on UCF101. 
These accuracies from the state-of-the art action recognition methods are always obtained with the of inclusion of several other modalities, such as optical flow, as well as on massive supervised datasets like ImageNet~\cite{Deng2009}.
With the obtained results, we show some promising directions in self-supervised learning for video data, which contribute towards self-supervised deep networks that could be alternatives to fully supervised or semi-supervised networks in the future.
\begin{table}[t]
\scriptsize
\begin{center}
\begin{tabular}{|l|c|c|c|c|}
\hline
Method 			  & UCF101-split1 	& HMDB51-split1\\ \hline \hline
DrLim~\cite{Hadsell2006}  &     45.7 		&   16.3  	 \\ \hline
TempCoh~\cite{Mobahi2009} &     45.4 		&   15.9  	  \\ \hline 
Obj. Patch~\cite{Wang2015}&     40.7 		&   15.6   	  \\ \hline
Seq. Ver.~\cite{Misra2016}&     50.9 		&   19.8   	  \\ \hline 
Our - Stack-of-Diff.	  &     \textbf{60.3}	&      \textbf{32.5} 	  \\ \hline  \hline
Rand weights - Stack-of-Diff. &     51.3		& 28.3 \\ \hline
ImageNet weights - Stack-of-Diff. &     70.1		& 40.8 \\ \hline
\end{tabular}
\end{center}
\caption{Comparing with other state-of-the-art self-supervised learning methods for action classification using UCF101 and HMDB51 datasets.}
\label{tbl.soa}
\end{table}

\section{Conclusion}
\label{sec.conclusion}
We present odd-one-out networks (O3N), a new way to learn visual features for videos without using category level annotations.
During feature learning, our O3N learns to do analogical reasoning about the input data leading to better generalizable features.
Learned features are fine-tuned for action classification and obtained 60\% classification accuracy on UCF101 dataset without resorting to external information or models such as pre-trained networks, or optical flow features. 
Similarly, we outperform previous-state-of-the-art results on self-supervised learning for action classification on HMDB51 dataset by more than 12\%.
Our O3N can be applied over different kinds of temporal encoders. 
We experimented using three video-clip encoders showing consistent improvements across all of them.
In future, we aim to use our odd-one-out network to learn features for images and videos jointly in self-supervised manner.

\small{
\noindent
\textbf{Acknowledgement:} This research was supported by the Australian Research Council (ARC) through the Centre of Excellence for
Robotic Vision (CE140100016) and was undertaken on the NCI National Facility in Canberra, Australia. Hakan Bilen is supported by the ERC Starting Grant Integrated and Detailed Image Understanding (EP/L024683/1).}

{\small
\bibliographystyle{ieee}
\bibliography{main}
}

\end{document}